\documentclass[10pt,twocolumn,letterpaper]{article}

\usepackage[pagenumbers]{cvpr}      

\usepackage[dvipsnames]{xcolor}
\newcommand{\red}[1]{{\color{red}#1}}

\definecolor{cvprblue}{rgb}{0.21,0.49,0.74}
\usepackage[pagebackref,breaklinks,colorlinks,citecolor=cvprblue]{hyperref}
\usepackage{lipsum}
\usepackage{multirow}
\usepackage{etoc}
\usepackage{siunitx}

\newcommand{\simabb}[0]{GenH2R-Sim~}
\newcommand{\simabbns}[0]{GenH2R-Sim}

\newcommand{\revise}[1]{\textcolor{black}{#1}}
\usepackage{tabularx,ragged2e}
\def\paperID{5897} 
\def\confName{CVPR}
\def\confYear{2024}

\title{GenH2R: Learning Generalizable Human-to-Robot Handover\\via Scalable Simulation, Demonstration, and Imitation}

\author{
Zifan Wang\textsuperscript{\textasteriskcentered1,3}~~~
Junyu Chen\textsuperscript{\textasteriskcentered1,3}~~~
Ziqing Chen\textsuperscript{1}~~~
Pengwei Xie\textsuperscript{1}~~~
Rui Chen\textsuperscript{1}~~~
Li Yi\textsuperscript{\textdagger1,2,3}
\smallskip\\
\textsuperscript{1}Tsinghua University~~
\textsuperscript{2}Shanghai Artificial Intelligence Laboratory~~
\textsuperscript{3}Shanghai Qi Zhi Institute
\\
\href{https://GenH2R.github.io}{\textcolor{magenta}{https://GenH2R.github.io}}
}

\begin{document}
\maketitle

\renewcommand{\thefootnote}{\fnsymbol{footnote}}
\footnotetext[1]{Equal contribution with the order determined by rolling dice.}
\footnotetext[2]{Corresponding author.}

\etocdepthtag.toc{mtchapter}

\begin{abstract}
This paper presents GenH2R, a framework for learning generalizable vision-based human-to-robot (H2R) handover skills. The goal is to equip robots with the ability to reliably receive objects with unseen geometry handed over by humans in various complex trajectories. We acquire such generalizability by learning H2R handover at scale with a comprehensive solution including procedural simulation assets creation, automated demonstration generation, and effective imitation learning. We leverage large-scale 3D model repositories, dexterous grasp generation methods, and curve-based 3D animation to create an H2R handover simulation environment named \simabbns, surpassing the number of scenes in existing simulators by three orders of magnitude. We further introduce a distillation-friendly demonstration generation method that automatically generates a million high-quality demonstrations suitable for learning. Finally, we present a 4D imitation learning method augmented by a future forecasting objective to distill demonstrations into a visuo-motor handover policy. Experimental evaluations in both simulators and the real world demonstrate significant improvements (at least +10\% success rate) over baselines in all cases.
\end{abstract}    
\section{Introduction}
\label{sec:intro}

\begin{figure}[t]
    \centering
    \includegraphics[width=0.95\linewidth]{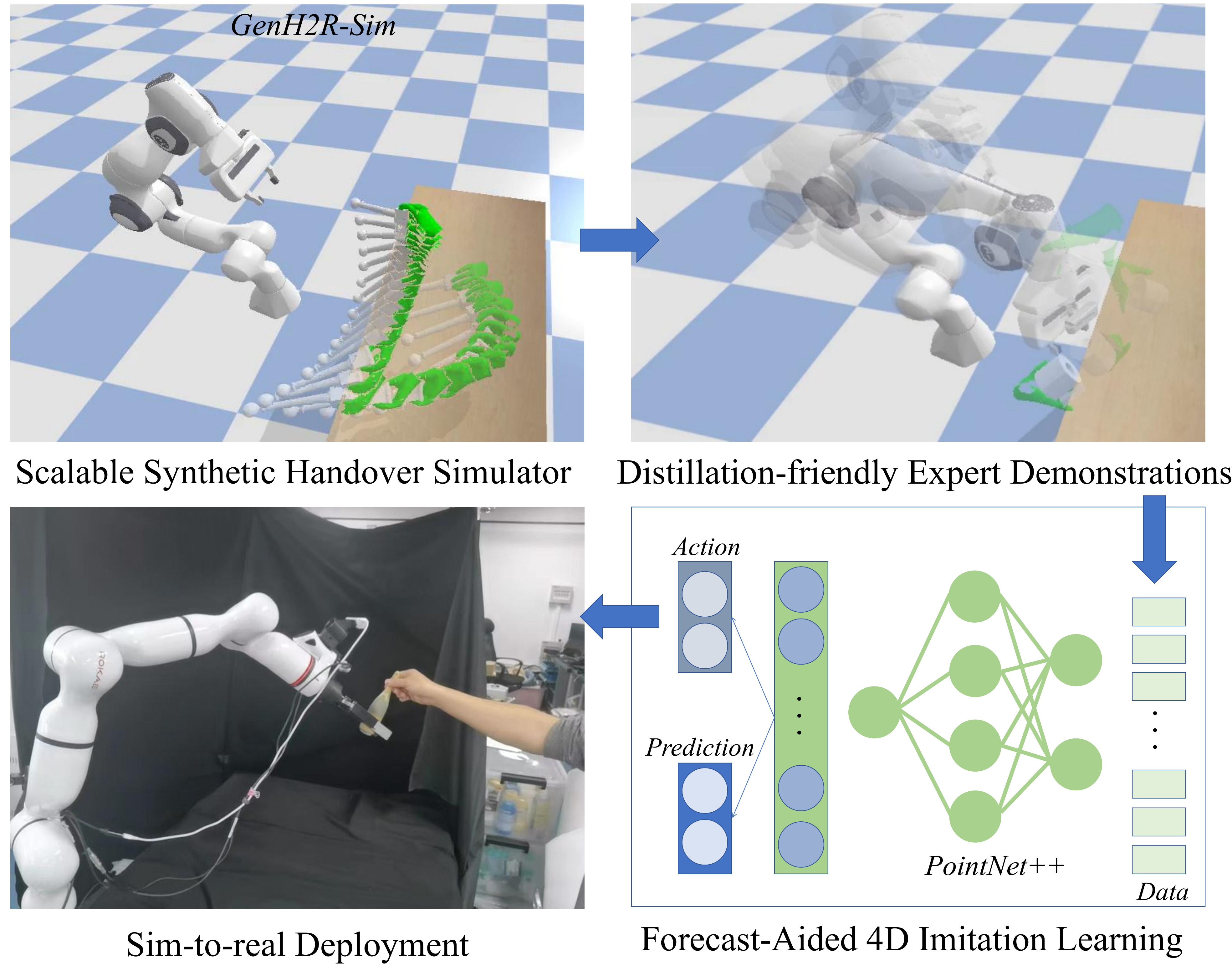}
    \caption{\textbf{The overview of GenH2R.} We introduce a framework for learning generalizable vision-based human-to-robot handover via scalable synthetic simulation, distillation-friendly expert demonstration generation, and a forecast-aided 4D imitation learning method. Our models demonstrate strong generalization capabilities to real datasets and can be deployed to a real robot.}
    \label{fig:teaser}
    \vspace{-2em}
\end{figure}

The embodied AI research community has long been driven by the goal of empowering robots to interact and collaborate with humans. A crucial aspect of this pursuit is equipping robots with the capability to reliably receive arbitrarily moving objects of varying geometry handed over by humans, based on dynamic visual observations. This human-to-robot (H2R) handover ability allows robots to seamlessly collaborate with humans across a wide range of tasks, including cooking, room tidying, and furniture assembly.

However, compared to learning human-free robot manipulation skills, the progress in scalably learning H2R handover that can generalize to various objects and versatile human behaviors has lagged due to its unique challenges. Training robots to interact with humans in real-world scenarios entails increased risks and expenses, rendering it inherently non-scalable. Therefore, it is demanded to simulate human behaviors and train robots in simulated environments prior to real-world deployment. However, creating a substantial number of assets for humans handing over objects poses a significant challenge. In a recent study~\cite{chao2022handoversim} that employed motion capturing to drive virtual humans in a simulator, only 1000 unique human hand motion trajectories were provided for handing over 20 objects. Limited object geometry and human motion assets can hardly capture the complexities of the real world. Besides, the challenge extends to the demonstration side. The success of large language model~\cite{brown2020language,zhang2022opt,openai2023gpt4} has suggested a recipe for scaling up learning through modeling large-scale training data. Nevertheless, collecting robot demonstrations receiving objects from real humans is very costly and unscalable. How to scale up the number of demonstrations while ensuring effective learning poses additional challenges.

In this work, we aim to learn generalizable H2R handover at scale by tackling the above challenges. We present a comprehensive solution that scales up both the assets and demonstrations and effectively learns a closed-loop visuomotor policy through a novel imitation learning algorithm. 

Specifically, to scale up geometry and motion assets depicting humans handing over various objects, we leverage large-scale 3D model repositories~\cite{chang2015shapenet,eppner2021acronym}, dexterous grasp generation methods~\cite{wang2023dexgraspnet}, and curve-based 3D animation. This enables us to procedurally generate millions of handover scenes, forming an environment named \simabb to support generalizable H2R handover learning. \simabb surpasses HandoverSim~\cite{chao2022handoversim}, an existing H2R simulator, in both scene quantity (by three orders of magnitude) and unique object involvement (by two orders of magnitude). In addition, scenes in \simabb go beyond a straightforward giving and then receiving and cover cases when humans might keep transforming the object in a large range during the entire H2R handover process. This allows for studying complex behaviors such as humans hesitating before handing over.

To scale up robot demonstrations, we draw inspiration from the Task and Motion Planning (TAMP)~\cite{garrett2021integrated} literature and propose to automatically generate demonstrations with grasp and motion planning using privileged human motion and object state information. There are some straightforward ways to achieve this goal, such as using the privileged human handover destination information to plan a smooth demonstration. However, the problem is more challenging than it seems since the generated demonstrations need to be suitable for distilling into a visuomotor policy. We identify the vision-action correlation between visual observations and planned actions as the crucial factor influencing distillability and point out that due to the constraints of robot arm morphology one can easily generate observation-irrelevant actions and thus harm distillation. To tackle this challenge, we present a distillation-friendly demonstration generation method that sparsely samples handover animations for landmark states and periodically replans grasp and motion based on privileged future landmarks.

Finally, to distill the above demonstrations into a visuomotor policy, we utilize point cloud input for its richer geometric information and smaller sim-vs-real gap compared to images. We propose a 4D imitation learning method that factors the sequential point cloud observations into geometry and motion parts, facilitating policy learning by better revealing the current scene state. Furthermore, the imitation objective is augmented by a forecasting objective which predicts the future motion of the handover object. Since our demonstrating actions are generated based on future landmarks, the forecasting objective can help further exploit the vision-action correlation.

We evaluate our learned policy in simulators (HandoverSim and our own \simabbns) and the real world. Remarkably, without any mocap assets or real-world demonstrations, our method achieves significantly better performance compared to baselines across all settings (at least \textbf{+10\%} success rate). Our experiments highlight that the scaling-up efforts bring substantial improvement in policy generalizability to novel geometry and complex motion. Furthermore, these efforts greatly facilitate skill transfer to real robotic systems.

In summary, the key contribution of this paper is a novel framework scaling up the learning of H2R handover with the following three components: i) a simulation environment named \simabb consists of millions of human handover animations for generalizable H2R handover learning, ii) an empirically validated automatic robot demonstration generation pipeline for vision-based closed-loop control, iii) a forecast-aided 4D imitation learning method effective in distilling the large-scale demonstrations.

\begin{center}
\begin{figure*}[t]
  \centering
  \includegraphics[width=2\columnwidth]{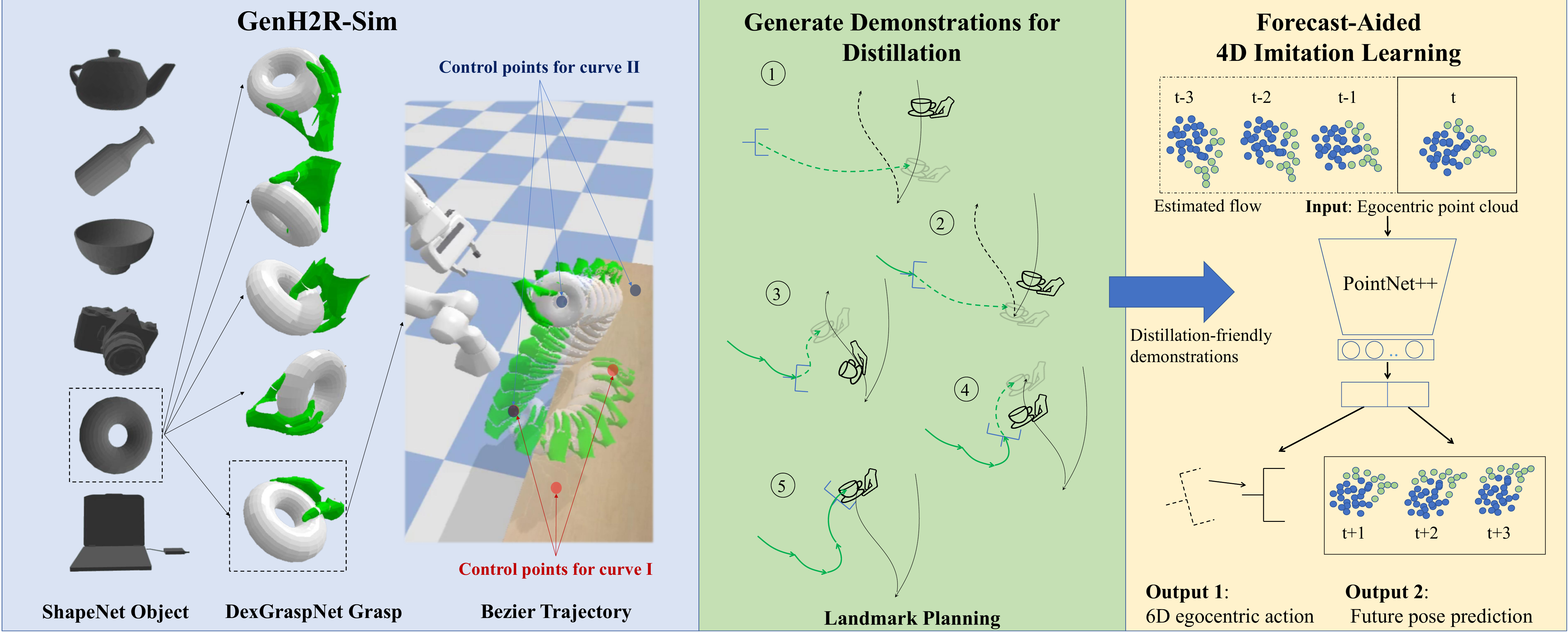}
  \caption{\textbf{The overview of our framework. } First, we propose a new simulation environment named \simabbns, featuring large-scale synthetic datasets with diversity in object geometry, grasp poses, and complex trajectories. Second, other than destination planning (move straight toward the final position) and dense planning (replan at each step), we propose a distillation-friendly demonstration generation method—landmark planning, predicting landmarks on the trajectory (as indicated by the dashed object above) and replanning based on those landmarks. Thirdly, our Forecast-aided 4D Imitation Learning leverages past flow information, and the forecasting objective enhances the exploitation of vision-action correlation.}
\label{fig:overview}
\vspace{-0.4cm}
\end{figure*}   
\end{center}
\section{Related Work}
\label{sec:related}

\subsection{Human-to-Robot Handovers}

Recently, significant progress in human-robot handovers \cite{ortenzi2021object,rosenberger2020object,corsini2022nonlinear} has been observed, driven by the increasing popularity of human-robot interaction~\cite{sheridan2016human,bartneck2020human} and the emergence of extensive datasets \cite{hasson2019learning,chao2021dexycb,ye2021h2o,fan2023arctic,liu2022hoi4d,brahmbhatt2020contactpose} capturing hand-object interactions. Some traditional methods~\cite{bicchi2000robotic,bohg2013data} require 3D object  models and struggle to handle unseen objects. One possible way is to consider grasping and dynamic motion planning~\cite{zhang2023flexible,yang2021reactive,fang2023anygrasp,marturi2019dynamic}. However, these methods often exhibit constrained motions and perform poorly on large-scale datasets. HandoverSim~\cite{chao2022handoversim}, a physics-simulated environment, introduced a new simulation benchmark for human-to-robot object handovers. Leveraging DexYCB~\cite{chao2021dexycb}, a dataset of human grasping objects and performing handover attempts, this environment allows training learning-based handover policies such as~\cite{christen2023learning}. However, it lacks large-scale and diverse handover scenes, which limits generalizable handovers. At the same time, SynH2R~\cite{christen2023synh2r} proposes to use synthetic data but makes limited progress. Building on this, we propose \simabbns~, aiming to benchmark generalizable handover.

\subsection{Scaling Up Robot Demonstrations}
For robot learning, scaling up data collection for manipulation skills has spurred extensive research. Approaches include leveraging large language models~\cite{ha2023scaling} or  hardware capabilities~\cite{song2020grasping}, utilizing non-robotics datasets~\cite{grauman2022ego4d}, and employing trial-and-error explorations~\cite{fu2020d4rl}. As depicted in ~\cite{ha2023scaling}, one of the challenges is scaling up robot-complete data. A popular line of research scales up demonstration generation via Task and Motion Planning~\cite{garrett2021integrated, mcdonald2022guided, dalal2023imitating}. These works usually focus on fairly static scenes without active motion or object and task variety~\cite{wang2023robogen, wang2023gensim} while our method extends to dynamic H2R handover by considering how to interpret human behavior and generate demonstrations easy to be distilled by closed-loop visuo-motor policy.

\subsection{Offline Learning from Demonstrations}
Imitation Learning (IL) represents a methodology for training embodied agents in manipulation tasks by utilizing expert demonstrations. The commonly used Behavior Cloning (BC) \cite{pomerleau1988alvinn} strategy directly trains the policy to imitate expert actions in a supervised learning manner. Despite its simplicity, this approach has demonstrated remarkable effectiveness in robotic manipulation~\cite{mandlekar2021matters,finn2017one,billard2008survey,zhang2018deep} especially when combined with a substantial number of high-quality demonstrations~\cite{jang2022bc,ebert2021bridge}. Inspired by these works, we adopt an imitation learning paradigm, focusing on how to leverage spatial-temporal perception and future forecasting to better consume our distillation-friendly demonstrations.

\section{Method}

\subsection{Overview}
\label{subsec:overview}



\revise{For the generalizable H2R handover task, we introduce GenH2R, a framework designed to learn control policies, specifically 6D control actions for the robot gripper, using segmented point cloud data captured from an egocentric camera.}
We describe our method for synthesizing human handover animations in Section \ref{subsec:sim}, generating expert demonstrations in Section \ref{subsec:demo}, and distilling demonstrations to 4D vision-based neural networks by imitation learning in Section \ref{subsec:4dnetwork}, as the pipeline depicted in Figure \ref{fig:overview}.

\subsection{\simabbns}
\label{subsec:sim}
The size and quality of human-object datasets in simulators play a crucial role in generating high-quality handover demonstrations and training reliable policies for handover scenarios. The recent handover simulator, HandoverSim~\cite{chao2022handoversim}, utilizes the DexYCB~\cite{chao2021dexycb} dataset, which captures real-world human grasping objects in a limited manner, comprising only 1000 scenes with 20 distinct objects. In the real world, scenarios can be more complex and may involve intricate trajectories and poses beyond those in DexYCB.

To address these limitations, we introduce a new environment, \simabbns, to overcome these deficiencies and facilitate generalizable handovers. To diversify geometry and motion assets depicting humans handing over various objects, we focus on two primary aspects: the hand grasping pose and the hand-object moving trajectory within a scene.

In aspects of grasping poses, DexGraspNet~\cite{wang2022dexgraspnet} has made significant contributions by employing optimization techniques to generate a substantial dataset of human hand grasp poses. We utilize this method to generate approximately 1,000,000 grasp poses for 3,266 different objects sourced from Shapenet~\cite{chang2015shapenet}. These objects span a wide range of categories, from larger items like computers to smaller ones like mobile phones, covering most sizes and shapes encountered in real-life handovers.

In aspects of hand-object moving trajectories, we propose to use B\'ezier curves, which are one class of smooth curves determined by several control points, to generate complex yet smooth-transiting motion trajectories. We use multiple B\'ezier curves to model different stages of the motion, and link the ends of these curves to create a seamless track. We can generate scenes matching various scenarios of different complexity in the real world by adjusting the distribution of control points of the trajectory and the speed of the human hand. To enhance the trajectory's realism, we incorporate consistent object rotations, which also enhances the importance of choosing the appropriate grasp for the robotic arm. Since we can always attach a new segment of motion at the end of the current motion and the duration is much longer than DexYCB scenes, the destination of the hand-object is not a significant factor, so we just randomly select a point within the reach of the robotic arm.

We do not guarantee that every item in the dataset we generate perfectly mimics the human-like characteristics of real-world data, but our approach ensures a significantly higher degree of domain randomization and provides greater diversity in terms of geometry and motion. Given the challenges in scaling up real-world motion capture datasets, we opt for a large-scale synthetic dataset for our handover simulations. Our key insight is that for both demonstrations and policy learning, having a substantial amount of synthetic data is more beneficial than relying on a small-scale real-world dataset.

GenH2R-Sim follows the setup of HandoverSim, which consists of a Panda 7DoF robotic arm with a gripper and a wrist-mounted RGB-D camera, and a simulated human hand. Just like HandoverSim, we switch from the pre-handover kinematic phase to the handover dynamic phase when the object has been in contact with the gripper. HandoverSim is not adaptive to the robot's action and just loads and replays every frame of the data. To align with the real-world handover process more naturally in GenH2R-Sim, the simulated hand will stop from moving and wait for handover when the robot arm is close to the object.

\subsection{Generating Demonstrations for Distillation}
\label{subsec:demo}

\begin{figure}[t]
  \includegraphics[width=\columnwidth]{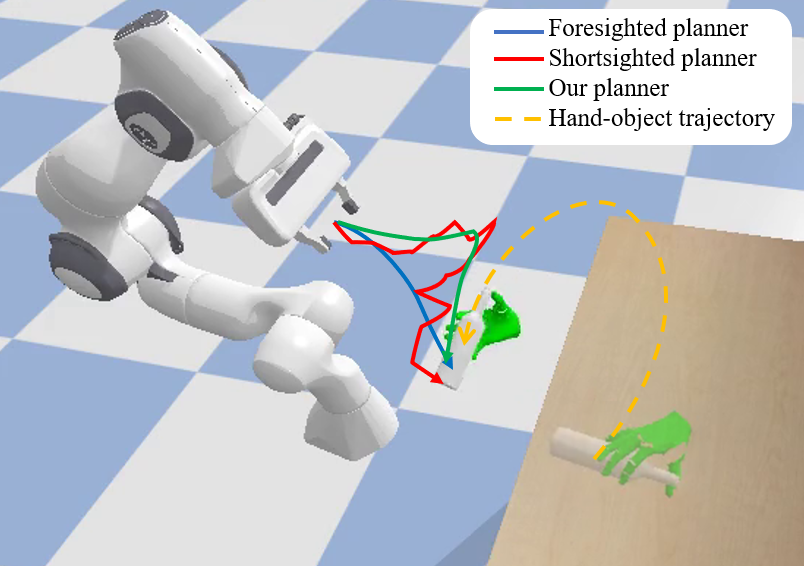}
  \caption{\textbf{Different demonstration generation methods for dynamic handover.} The orange curve shows the hand-object trajectory. The blue, red, and green curves show the example trajectories generated by the foresighted planner, the shortsighted planner, and our planner, respectively.}
\label{fig:two_strategies}
\vspace{-0.4cm}
\end{figure}

In this section, we address a key question in learning visuo-motor policy: how to efficiently generate robot demonstrations that incorporate paired vision-action data from successful task experiences. While distilling successful demonstrations into a single policy has proven effective for open-loop control tasks, the challenge lies in closed-loop visuo-motor control, where the quality of demonstrations becomes crucial for learning. Merely ensuring success is no longer sufficient. We present two examples of demonstration generation with different grasp and motion planning strategies as shown in Figure~\ref{fig:two_strategies}. In the first example, a foresighted planner generates smooth, short demonstrations based on the privileged destination end state of a human handover animation. Though efficient, the planned path does not align actions with the dynamic visual observations during the handover. Distilling such demonstrations requires accurately forecasting the end state of the human trajectory, which can be extremely challenging in complex handover cases. The second example involves a shortsighted planner that independently replans grasp and motion at each time step using privileged hand and object states. Due to robot morphology constraints and the multi-solution nature of common robot planners, smooth visual observations may correspond to unsmooth and multi-modal robot trajectories, increasing the difficulty of distillation. We emphasize the importance of distillability as a quality factor for handover demonstrations. An effective demonstration generation method must consider the vision-action correlation by jointly incorporating robot morphology and dynamic vision during grasp and motion planning.

Along this line, we base our method on the foresighted and shortsighted planner mentioned above to combine the advantages of both sides while encouraging the demonstration distillability. We first improve the shortsighted planner so that sequentially smooth visual observations result in smooth grasp and motion plans. Then we improve the handover efficiency by looking toward the future while guaranteeing the vision-action correlation. 

To be specific, we build our method based on the OMG planner~\cite{wang2020manipulation} for grasp and motion planning. This planner optimizes the grasp and motion path by considering the object's 6D pose and a set of candidate grasp poses. To support this optimization, we provide privileged knowledge that includes the object's 6D pose, candidate grasps generated through physics simulation~\cite{eppner2021acronym}, and human hand poses for filtering out invalid grasps. However, independently calling the OMG planner for each time step may result in unsmooth trajectories, as it is designed for static scenarios. To address this, we sequentially plan the grasp and motion based on the privileged knowledge by: 1) sorting grasps based on their pose distance to the robot end effector and attempting inverse kinematics (IK) starting from the nearest grasp until success; 2) initializing IK based on the robot arm pose from the previous time step; 3) invoking the OMG planner only when IK can be successfully solved. By prioritizing closer grasps, we encourage the object to remain within the field of view of a wrist camera, reducing visually irrelevant actions when the object is not visible. Additionally, enforcing IK smoothness improves the overall trajectory smoothness. As a result, the enhanced vision-action correlation dramatically improves the demonstration quality.

Our approach modifies the OMG planner for dynamic grasp and motion planning. However, densely replanning at each time step leads to inefficient and non-smooth zigzag demonstrations, which does not align with how humans receive objects. Humans anticipate dynamic scene changes before taking action. On the other hand, a highly foresighted planner that directly plans grasp and motion based on the end state of a human handover animation can disrupt the vision-action correlation. To strike a balance between these extremes, we propose an algorithm that sparsely samples handover animations for landmark states and periodically replans grasp and motion based on future landmarks. The key idea is to select landmarks strategically so that the planner only considers visually foreseeable futures. Specifically, let \(\xi=(\mathcal T_0,\mathcal T_1,\cdots,\mathcal T_{T-1})\) represent an object trajectory, where \(\mathcal T_t\in \mathbb{SE}(3)\) denotes the object pose in the \(t\)-th frame within the world coordinate system. Based on all the object trajectories in the training set, we train an object pose forecasting network which consumes past and current object poses $(\mathcal T_0,\mathcal T_1,\cdots,\mathcal T_{t})$ for each time step $t$ within each trajectory and forecasts the object poses $(\mathcal T_{t+1},\mathcal T_{t+2},\cdots,\mathcal T_{t+N})$ in future $N$ steps. By thresholding the forecasting error corresponding to each time step, we identify a set of endpoints where past observations cannot forecast the future very well and partition the complete trajectory \(\xi\) to several segments using endpoints \(0=l_0<l_1<\cdots<l_k=T\). Within each segment, we assume the ability to predict the future object pose based on historical information. We then denote \(P\in\mathbb{N}\) as the hyperparameter determining the replaning period. For each planning frame \(t=0, P, 2P, \cdots\), suppose the next endpoint is \(l_{i+1}\), \ie, \(l_i\le t<l_{i+1}\). Then we will plan based on the object pose at frame \(\hat t=\min(t+P, l_{i+1})\), which serves as a landmark. Note here planning is based on the future states but avoids bypassing the sharply transitioning points where human motion becomes unpredictable. Also worth mentioning, densely planning is a special case of our method, and landmark planning is a full version.

\subsection{Forecast-Aided 4D Imitation Learning}
\label{subsec:4dnetwork}
Traditional methods for human-to-robot handover face challenges in gaining insights into dynamic scene perception. Approaches based on motion planning~\cite{wang:rss2020} often emphasize robot morphology and lack dynamic vision perception. They struggle to capture long-horizon information, mainly focusing on the current frame and failing to predict the future. Reinforcement Learning methods~\cite{wang2022goal,christen2023learning}, while powerful, require extensive training and may train unstably across different scenarios. To enhance the vision-action correlation and establish an efficient training paradigm, we introduce our forecast-aided 4D imitation learning approach.

In robot perception, the 4D point cloud serves as the common representation. 
In the $t$-th frame, we can define $M_t^i \in \mathbb{SE}(3)$ as the relative object pose between the current frame and the $i$-th frame in the egocentric view. While frame stacking is a straightforward approach, it struggles to capture both motion and geometry effectively. Inspired by recent 4D learning methods~\cite{dong2023nsm4d,teed2021raft}, we employ the Iterative Closest Point (ICP) registration algorithm~\cite{rusinkiewicz2001efficient} to efficiently compute transformation matrices $\{\hat{M}_t^{t-1}, \hat{M}_t^{t-2},\cdots, \hat{M}_t^{t-L_1}\}$ between the point cloud in the $t$-th frame and the point clouds in previous $L_1$ frames. Applying these transformation matrices to a specific point in the current frame yields its rough coordinates in previous frames. Then we incorporate this flow feature into 3D PointNet++~\cite{qi2017pointnet++} to encode a global spatial-temporal feature and use Multilayer Perceptron (MLP) to decode it into a 6D egocentric action. The loss function, denoted as $\mathcal{L}_{action}$, is computed as the L1 loss for aligning 3D points on the robot gripper as defined in \cite{li2018deepim}. We believe some sophisticated 4D backbones~\cite{fan2021point,wen2022point} are suitable for 4D understanding, but they are often not suitable for robotic tasks that require a fast reference speed. Our method strikes a balance between effectiveness and simplicity.

To enhance the responsiveness of our policy to human motion and extend the vision horizon into the future, we introduce an auxiliary task to predict the future motion $\{M_t^{t+1}, M_t^{t+2},\cdots, M_t^{t+L_2}\}$ of objects in the next $L_2$ frames. Using the ground truth object poses from trajectories, we compute the motion prediction loss for the $t$-th frame:

\begin{equation}
\mathcal{L}_{pred} = \sum_{i=t+1}^{t+L_2} \|\hat{M}_t^{i} - {M}_t^{i}\|
\end{equation}

In contrast to reinforcement learning, our imitation learning method requires only a few hours of training and achieves great generalizability through large-scale, high-quality demonstrations. 
We acquire vision-action pairs and ground truth object states from demonstrations, and then supervise our policy using the loss function $\mathcal{L} = \mathcal{L}_{action} + \lambda\mathcal{L}_{pred}$, where $\lambda$ serves as a weighting hyper-parameter to balance the losses. This efficient distillation paradigm empowers our policy to naturally approach objects with a forecasting intention and to effectively generalize to a wide range of unseen objects and motions.

\section{Experiments}

\textbf{Dataset} (1) HandoverSim~\cite{chao2022handoversim} contains 1000 real-world H2R handover scenes and 20 objects from DexYCB~\cite{chao2021dexycb}. We evaluate on the ``s0'' setup which contains 720 training and 144 testing scenes. Each handover motion has a duration of 3 seconds. Following the evaluation of Handover-Sim2real~\cite{christen2023learning}, we consider ``Sequential'' and ``Simultaneous'' settings. In ``s0 (Sequential)'', the robot is allowed to move when the hand reaches the handover location and remains static. In ``s0 (Simultaneous)'', the robot is allowed to move
from the beginning of the episode.
(2) \simabb contains 1,000,000 complex synthetic H2R handover scenes and 3266 objects. We evaluate the ``t0'' setup which contains 1,000,000 training and 3260 testing scenes. Each handover motion has a duration of 8s and will stop when the robot gripper is close to the object. To introduce more real-world handover scenes into \simabb for evaluation, we extract and clip the handover point cloud sequence from HOI4D~\cite{liu2022hoi4d}, a real-world mocap dataset. This additional setup is referred to as ``t1'', which only contains 1000 testing scenes for evaluation. 

\begin{table*}[t]
    \vspace{-1em}
    \centering
    \small
    \begin{tabularx}{1.0\linewidth}{cc|ccc|ccc|ccc|ccc}
    \cline{1-14}
    \multicolumn{2}{c|}{\multirow{2}{*}{}}                                    & \multicolumn{3}{c|}{s0 (Sequential)}         & \multicolumn{3}{c|}{s0 (Simultaneous)}            & \multicolumn{3}{c|}{t0}                      & \multicolumn{3}{c}{t1}                      \\
    \multicolumn{2}{c|}{}                                                     & S          & T   &  AS        & S          & T  &  AS            & S         & T   &  AS  & S          & T  &  AS          \\ \cline{1-14}

     & OMG Planner$\dagger$~\cite{wang:rss2020}& 62.50 &  8.31  &  22.5 & -  & - & - & - & - & - & - & - & -\\ \cline{1-14}

    \multicolumn{1}{c|}{\multirow{6}{*}{s0}} & GA-DDPG~\cite{wang2022goal}  & 50.00 &  \textbf{7.14} & 22.5 & 36.81 & \textbf{4.66} & 23.6 & 23.59 & 7.31 & 10.3 & 46.7 & \textbf{5.50} & 26.9\\
    
    \multicolumn{1}{c|}{\multirow{2}{*}{train on}}   &Handover-Sim2real~\cite{christen2023learning}& 75.23 &    7.74  &  \textbf{30.4}  &  68.75  & 6.23 & 35.8 & 29.17 & 6.29 & 15.0 & 52.40 & 7.09 & 23.8\\
    
    \multicolumn{1}{c|}{}  &Handover-Sim2real\red{*}~\cite{christen2023learning} & 64.35 &  7.61  &  26.7  &  25.69  & 5.43 & 15.0 & 28.56 & 4.73 & 17.9 & 30.60 & 5.98 & 16.5\\

    \cline{2-14}
    
    \multicolumn{1}{c|}{} &  Destination Planning   &  74.31 &  9.01 & 22.8 & 76.16 & 6.98 & 35.2 & 25.68 & 5.96 & 14.1 & 48.4 & 8.94 & 15.1\\
    
    \multicolumn{1}{c|}{} & Dense Planning & 74.77 & 9.54 & 19.8 & 75.45 & 7.32 &  33.0 &  27.30 & 6.26 & 14.1 & 52.3 & 9.24 & 15.1 \\

    \multicolumn{1}{c|}{} & Landmark Planning & 77.78 & 9.24 & 22.3 & 79.17 & 7.26 & 34.9 & 29.63 & 6.23 & 15.4 & 54.2 & 9.02 & 16.6\\  
    \cline{1-14}
    \cline{1-14}

    \multicolumn{1}{c|}{\multirow{6}{*}{t0}} & {GA-DDPG}~\cite{wang2022goal} & 54.76 & 7.26 & 24.2  & 44.68 & 5.30 & 26.5  & 24.05 & 4.70 & 15.3  & 25.50 & 5.86 & 14.1\\
    
    \multicolumn{1}{c|}{\multirow{2}{*}{train on}}  & {Handover-Sim2real}~\cite{christen2023learning}    &  65.97 &  7.18  & 29.5 & 62.50 & 6.04 & 33.5 & 33.71 & 5.91 & 18.4 & 47.10 & 6.35 & 24.1\\

    \multicolumn{1}{c|}{}  & {Handover-Sim2real\textcolor{red}{*}}~\cite{christen2023learning}    & 63.55  &  7.58 & 26.5 & 38.89 & 5.29 & 23.1 & 33.31 & \textbf{4.64} & 21.4 & 33.35 & 5.81 & 18.4 \\
    \cline{2-14}

    \multicolumn{1}{c|}{}  & Destination Planning   & 0.93 &  12.80 & 0.01 & 6.48 & 12.41 & 0.3 & 5.96 & 8.81 & 1.9 & 1.60 & 12.03 & 0.1\\
    
    
    \multicolumn{1}{c|}{} & Dense Planning  & 81.48 & 9.51 & 21.9 & 84.95 & 7.45 & 36.3 & 38.04 & 7.16 & 17.1 
    & 57.90 &  8.85 & 18.4\\

    \multicolumn{1}{c|}{} & Landmark Planning & \textbf{86.57} & 8.81 & 28.0 & \textbf{85.65}  & 6.58 & \textbf{42.8} & \textbf{41.43} &  6.01 &\textbf{22.3} &\textbf{68.33} &  7.70 & \textbf{27.9}\\
    
    
    \cline{1-14}
    \end{tabularx}
     \caption{\textbf{Evaluating on different benchmarks. } We compare our method against baselines from the test set of HandoverSim \cite{chao2022handoversim} benchmark (``s0 (sequential)'' and ``s0 (simultaneous)'') and our \simabbns~benchmark (``t0'' and ``t1''). We use the best-pretrained models from the repositories of GA-DDPG~\cite{wang2022goal} and Handover-Sim2real~\cite{christen2023learning} for evaluation. The results for our method are averaged across 3 random seeds. Note that S means success rate(\%). T means time(s). AS means average success(\%) as defined in Equation \ref{equ:ap}. $\dagger$: This method \cite{wang:rss2020} is evaluated with ground-truth states and cannot handle dynamic handover like ``s0 (Simultaneous)'', ``t0'' and ``t1''.\textcolor{red}{*}: \revise{We reproduce the results of HandoverSim2real in the true simultaneous setting as detailed in Section \ref{exp:benchmark} to make a fair comparison.}
    }
     \label{tab:main_exp}
    \vspace{-1em}
    \end{table*}

\noindent\textbf{Metrics} We adhere to the HandoverSim evaluation protocol. A successful handover involves grasping the object from the human hand and moving it to a designated location. Failure cases involve hand contact, object drop, and timeout ($T_{max}=13s$). We report the successful rate and the execution time. Given that some policies prioritize success over speed, potentially wasting considerable human time, and others prioritize speed without considering success, we aim to evaluate both success rate and completion efficiency. To achieve this, we introduce AS (Average Success), akin to AP (Average Precision):
\begin{equation}
\text{AS} = \int_0^1 \text{Success}(t) \, \mathrm{d}t
\label{equ:ap}
\end{equation}
where $\text{Success}(t)$ is success rate considering only successful cases within $t \cdot T_{\text{max}}$. This method can better evaluate success-time relations which is more suitable in our handover scenarios.

\subsection{Evaluating on Different Benchmarks}
\label{exp:benchmark}
\textbf{Setup} We have 2 training sets: small-scale real-world ``s0'' from HandoverSim and large-scale synthetic ``t0'' from our \simabbns.~Evaluation is conducted on four testing sets: ``s0 (Sequential)''/``s0 (Simultaneous)'' from HandoverSim and ``t0''/``t1'' from our \simabbns.  We conduct experiments on our forecast-aid 4D imitation learning from different demonstration strategies including destination planning, dense planning, and landmark planning. \revise{As discussed in Section \ref{subsec:demo}, destination planning denotes the foresighted planner, dense planning denotes the improved shortsighted planner and landmark planning is our proposed method.} 

\noindent\textbf{Baselines} We compare our methods with Handover-Sim2real\footnote{
\revise{Our approach strictly adheres to the simultaneous setting defined in the paper of HandoverSim and HandoverSim2real:  the robot moves from the beginning of the handover episode. However, it's noteworthy that HandoverSim2real manually makes their policy hold still in the first 1.5 seconds in the code implementation, deviating from the simultaneous setting definition. To ensure a fair comparison, we reproduce their results in the true simultaneous setting.}},
the state-of-the-art method in HandoverSim. We additionally compare GA-DDPG which is designed for grasping objects, and OMG Planner.

\noindent\textbf{Results on different datasets} As depicted in Table \ref{tab:main_exp}, our method trained on ``t0'' outperform all methods trained on ``s0''  by a large margin. Compared with Handover-Sim2real trained on ``s0'', our landmark planning method trained on ``t0'' exhibits 11.34\%, 16.90\%, 12.26\%, and 15.93\% increase in the success rate across the four testing sets. 
Moreover, compared with our landmark planning method trained on ``s0'', the version trained on ``t0'' demonstrates notable improvements, achieving success rate increases of 8.79\%, 6.48\%, 11.80\%, and 14.13\% increase in the same testing sets. 
This underscores the importance of having a substantial amount of synthetic data for handover training in simulation, which is more beneficial than only relying on a small-scale real-world dataset. Our \simabbns, with its large-scale complex human hand behavior, generalizes effectively to real-world scenarios such as ``s0'' in DexYCB and ``t1'' in HOI4D.




\noindent\textbf{Results for different methods} We can compare our methods with the baseline HandoverSim2real within the same training set in different benchmarks. When trained on ``s0'', our landmark planning method demonstrates improvements of  2.55\%, 10.42\%, 0.46\%, and 1.8\% (13.43\%, 53.48\%, 1.07\%, and 23.60\% in our reproduced version) across the 4 test sets. Similarly, When trained on ``t0'', our landmark planning method gives substantial improvements of 20.78\%, 23.15\%, 7.72\%, and 21.23\% (23.02\%, 46.76\%, 8.12\%, and 34.98\% in our reproduced version).
The last 3 benchmarks (``s0''(simultaneous), ``t0'', and ``t1'') closely resemble real-world scenarios. They greatly demonstrate the effectiveness of our pipeline from distillation-friendly demonstrations to forecast-aided 4D imitation learning, which is capable of handling dynamic robot perception in complex handover scenarios. 
We also show visualizations on different methods in Figure ~\ref{fig:real_world} (a)(b).

\noindent\textbf{Results for different Demonstrations} 
Trained on ``s0'' which consists of relatively simple trajectories, demonstrations based on destination planning can offer a rudimentary cue for downstream visuo-motor policy. However, when trained on ``t0'' this strategy may lose focus on the object, leading to a failure in maintaining vision-action correlation and providing minimal gains for vision-friendly learning. There is a significant 73.38\% / 69.68\% decrease in success rate in the ``s0'' setting.
Additionally, distillation from landmark planning slightly outperforms dense planning in success rate and completes the handover process more quickly in all benchmarks. While dense planning can sustain the success rate to some extent, it slows down the agent and may result in unnatural approaches to objects. To jointly consider the time efficiency and the success rate, we compare the Average Success in methods distilled from these two strategies and find that landmark planning is a more efficient and generalizable approach.
For instance, when trained on ``t0'', landmark planning exhibits significant improvements of 6.1\%, 6.5\%, 5.2\%, and 9.5\% across the four testing sets.



\subsection {Evaluating on different Dataset Scales}
We have proved the crucial role of large-scale datasets in handover generalization in Section \ref{exp:benchmark}. We can also reveal it by scaling down the usage of ``t0'' in \simabbns~which contains 1,000,000 training scenes. 
With 10\% data utilization, we observe a 5.93\% drop in the success rate on the unseen ``t1'' test set. 
This result proves the significance of the dataset scale in our imitation learning method. Thanks to our large-scale data and efficient demonstration generation pipeline, concerns about limited datasets hindering generalization are alleviated.






\subsection{Ablation Study}

\begin{table}[t]
\centering
\setlength{\tabcolsep}{3mm}
\small
\begin{tabularx}{1.0\linewidth}{cccc}
\hline
\textbf{Methods}              & ~~~~~~S~~~~~~ & ~~~~~~T~~~~~~ &  ~~~~~~AS~~~~~~\\ \hline
w/o Flow            &     31.66        &   \textbf{5.67}   &  17.9\\ 
w/o Prediction       &  39.18     &     6.11        &      20.7\\
w/o Flow \& Prediction   &    37.04       &   5.93 &     20.1\\ 

Ours                 &       \textbf{41.43}      &    6.01  &   \textbf{22.3}\\ \hline

\end{tabularx}
\vspace{-1mm}
 \caption{\textbf{Ablations on different modules.} ``w/o Flow'' means do not use flow information in the input. ``w/o Prediction'' means do not add prediction loss in the output.
}
 \vspace{-1em}
 \label{tab:ablation}
\end{table}
\label{subsec:ablation}

As shown in Table \ref{tab:ablation}, we prove the effectiveness of our well-designed 4D imitation learning method.
The absence of flow information results in a 9.77\% decrease (predicting without past information adversely affects the model performance). The absence of the prediction task leads to a 2.25\% decrease, and the absence of both components results in a 4.39\% decrease.
The results demonstrate the model obtains improved performance in leveraging flow information, particularly when tasked with predicting the future object pose. More ablations about our demonstration generation and imitation learning are detailed in the supplementary material.


\subsection{Real World Experiments}

\textbf{Sim-to-Real Transfer} In addition to simulation, we deploy the models trained in \simabb on a real robotic platform. Using point cloud input from the wrist-mounted camera, we employ the output 6D egocentric action to update the end effector's target position. A user study compares our method against Handover-Sim2real~\cite{christen2023learning}. The supplementary material provides further details.

\begin{figure*}[t]
    \includegraphics[width=0.9\linewidth]{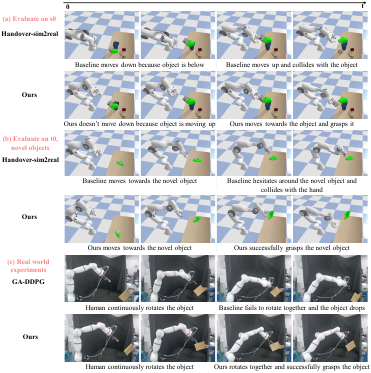}
    \vspace{-0.5em}
    \caption{\textbf{Qualitative results}. We in detail compare different methods in simulators and deploy them in the real-world platform.}
    \label{fig:real_world}
    \vspace{-1em}
\end{figure*}

\noindent \textbf{User Study} We recruited 6 users to compare our method (based on landmark planning) and Handover-Sim2real across 5 objects in 2 different settings. In the simple setting, users hand each object to the gripper without quick movements. In the complex setting, users execute a relatively long and quick trajectory. The results are reported in Table \ref{tab:real_world}.
We observe that our model gets better performance in completing the handover process across various objects and scenarios. Figure \ref{fig:real_world}(c) shows examples of the real-world handover trials.

\begin{table}[t]
\centering
\small
\begin{tabularx}{1.0\linewidth}{ccc}
\hline
\textbf{Methods}              & ~~~\textbf{Simple Setting}~~~  & ~~\textbf{Complex Setting}~~~  \\ \hline
Handover-Sim2real & 56.7\% & 33.3\% \\
Ours & 90.0\% & 70.0\% \\ \hline

\end{tabularx}
\vspace{-1mm}
 \caption{\textbf{Sim-to-Real Experiments.} We report the success rate of our method and HandoverSim2real in 2 different settings. 
}
 \vspace{-0.6cm}
 \label{tab:real_world}
\end{table}
\section{Conclusion}


In this work, we present a novel framework GenH2R for scaling up the learning of human-to-robot handover. We introduce a new simulator GenH2R-Sim and generate a million human handover animations to facilitate generalizable H2R handover learning. We then propose a distillation-friendly demonstration generation method that automatically produces a million high-quality demonstrations suitable for learning. We further introduce a forecast-aided 4D imitation learning method for effective demonstration distillation. Our experiments demonstrate that scaling-up efforts result in substantial improvement of generalizability to novel geometry and complex motion, both in the simulator and the real world.
{
    \small
    \bibliographystyle{ieeenat_fullname}
    \bibliography{main}
}

\clearpage
\appendix

\twocolumn[\begin{center}
    \large\bfseries GenH2R: Learning Generalizable Human-to-Robot Handover
via Scalable Simulation, Demonstration, and Imitation Supplementary Material\par
    \end{center}]
    
\addcontentsline{toc}{section}{Appendix}

The supplementary material offers additional details on various aspects of the method and experiments. Refer to the table of contents below for an overview. Section~\ref{sec:A} provides additional details and clarifications on our methods. Sections~\ref{sec:B} and~\ref{sec:C} present supplementary experiments on baselines, along with additional quantitative and qualitative results in the simulation and real-world scenarios, respectively. Section~\ref{sec:D} discusses the limitations of our work and explores potential research directions for future human-to-robot handovers and human-robot interactions.

\etocdepthtag.toc{mtappendix}
\etocsettagdepth{mtchapter}{none}
\etocsettagdepth{mtappendix}{subsection}
{
  \hypersetup{
    linkcolor = black
  }
  \tableofcontents
}

\section{More Method Details}
\label{sec:A}
\begin{figure*}[t]
    \centering
    \includegraphics[width=\linewidth]{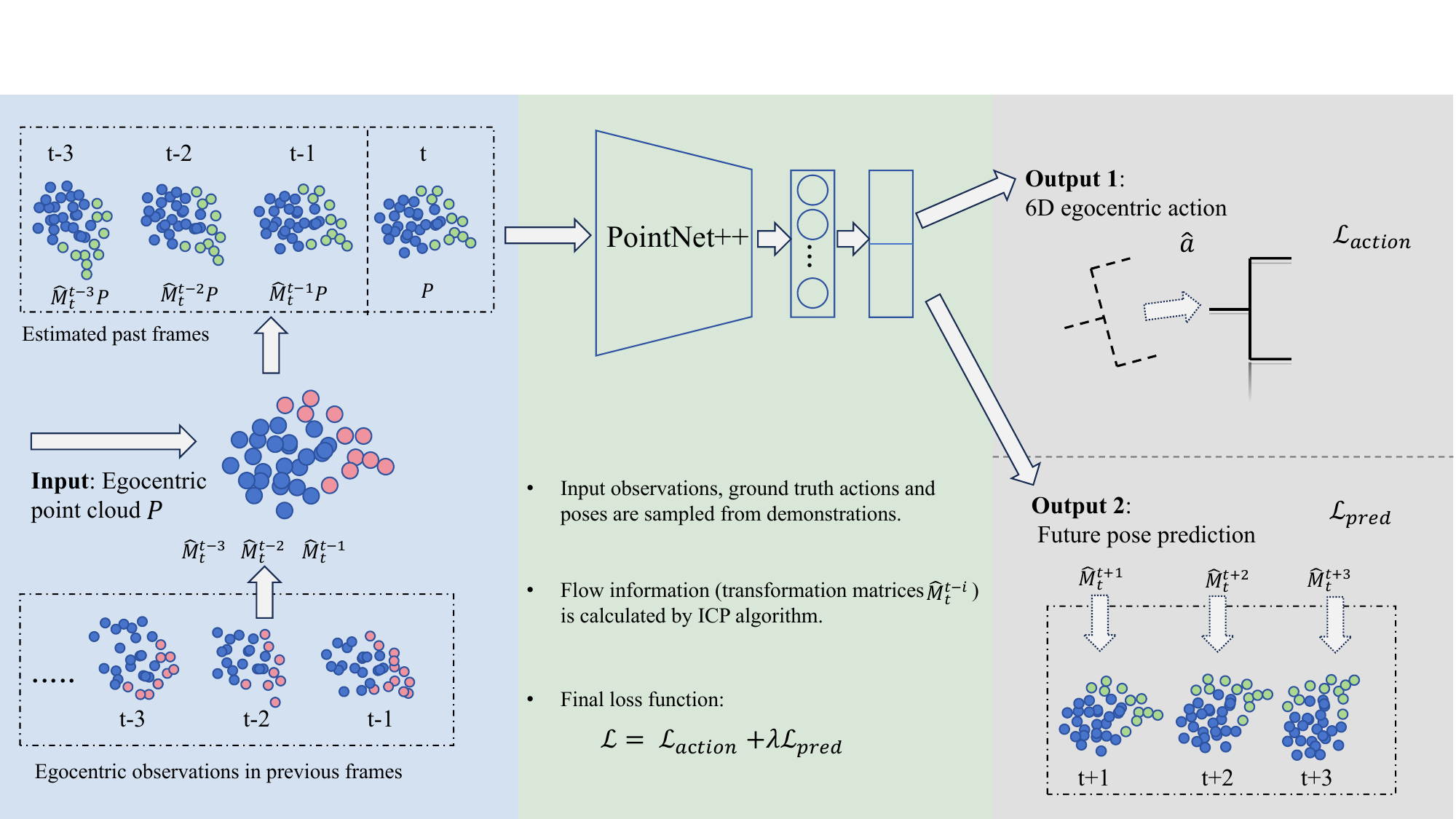}
    \caption{\textbf{Forecast-Aided 4D Imitation Learning Pipeline.} The network receives egocentric point cloud input and produces egocentric 6D actions as output. For each input point, we compute its past coordinates using flow information obtained through the  Iterative Closest Point algorithm. Subsequently, we employ PointNet++ to encode the processed point cloud into a low-dimensional global feature. The policy head decodes this feature into a 6D egocentric action, serving as the primary policy output. Simultaneously, the prediction head decodes the feature into future pose transformations, contributing to the auxiliary loss.}
    \label{fig:flow_pipeline}
\end{figure*}

\subsection{GenH2R-Sim}

In this section, we provide details on the generation of hand-object moving trajectories and our simulator \simabbns.

\subsubsection{Hand-Object Moving Trajectory Generation}

\textbf{Note:} The unit for all positions in the following paragraphs is meters.

In HandoverSim \cite{chao2022handoversim} and \simabbns, the robot arm's base is located at \((0.61, -0.50, 0.875)\), and the center of the table surface is at \((0.61, 0.28, H=0.92)\). To synthesize a hand-object moving trajectory, we start by generating the object's trajectory. 

For the translation part, we uniformly sample a starting point from the starting region \([0.3, 0.9]\times[0,0.2]\times[H+0.1, H+0.3]\). Subsequently, we sample several endpoints from the activity region \([0.1, 1.1]\times[-0.3, 0.1]\times[H+0.1, H+0.7]\) and employ B\'ezier curves to connect the starting point and the endpoints. For each B\'ezier curve, a key point is sampled from a Gaussian distribution centered at the midpoint with a standard variation of \(0.2\), and the translation speed along this curve is uniformly sampled from \([0.1,0.2]\) \si{\meter\per\second}.

For the rotation part, we uniformly sample a rotation \(R\in \mathrm{SO}(3)\) as the starting object orientation. When the object travels along a B\'ezier curve, we rotate the object about a random rotation axis with an angular speed uniformly sampled from \([0.5, 1]\) \si{\radian\per\second}.

After generating the object trajectory \(\xi=(\mathcal T_0,\mathcal T_1,\cdots,\mathcal T_{T-1})\), we correspondingly generate the hand pose trajectory \(\varsigma=\xi\circ T_{\text{object}}^{\text{hand}}\), where \(T_{\text{object}}^{\text{hand}}\) is the hand pose in the object reference frame.

\subsubsection{More Details about \simabbns}

In real-world handovers, in order to enhance stability, our motion typically stops when another person's hand is close to the object. We incorporate this characteristic into our simulator, making it reactive to the robot arm's motion. To be specific, suppose \(p\in\mathbb{R}^3\) is the current position of the gripper's tip, and \(Q\subset \mathbb{R}^3\) is the current object point cloud. If \(\min_{q\in Q}\|p-q\|\le 0.1\), the hand and object will stop moving, awaiting the robot arm to grasp the object.
We believe that this modification makes the cooperative handover process more realistic, transforming it from a simple chase-and-grasp game into a more authentic interaction.

It is important to note that we apply this modification exclusively in our simulation environment, GenH2R-Sim, for the benchmarks ``t0'' and ``t1''. We refrain from modifying it in HandoverSim~\cite{chao2022handoversim} for the benchmark ``s0'' to ensure a fair comparison with the exact results obtained by baseline methods.

\subsection{Generating Demonstrations for Distillation}
\subsubsection{Clarification of Different Methods}
We would like to provide further clarification regarding the terms used to describe various demonstration generation methods, as outlined in Table 3 of the manuscript. These terms include \textbf{destination planning}, \textbf{dense planning}, and \textbf{landmark planning}.

\textbf{Destination planning} refers to the foresighted planner discussed in Section 3.3, which plans directly to the object's destination at the beginning. While the generated demonstrations exhibit smoothness, they lack vision-action correlation in complex scenarios and are not distillation-friendly.

\textbf{Dense planning} represents the special case of our method, where the planner plans at each time step based on the current object position. Although the generated demonstrations ensure a strong vision-action correlation, they suffer from the zigzag issue, resulting in slower performance.

\textbf{Landmark planning} denotes our method, where the planner periodically plans based on the object's position at future landmarks.

\begin{table*}[t]
    \centering
    \small
    \begin{tabularx}{1.0\linewidth}{cc|ccc|ccc|ccc|ccc}
    \cline{1-14}
    \multicolumn{2}{c|}{\multirow{2}{*}{}}                                    & \multicolumn{3}{c|}{s0 (Sequential)}         & \multicolumn{3}{c|}{s0 (Simultaneous)}            & \multicolumn{3}{c|}{t0}                      & \multicolumn{3}{c}{t1}                      \\
    \multicolumn{2}{c|}{}                                                     & S          & T   &  AS        & S          & T  &  AS            & S         & T   &  AS  & S          & T  &  AS          \\ \cline{1-14}

     & OMG Planner$\dagger$~\cite{wang:rss2020}& 62.50 &  8.31  &  22.5 & -  & - & - & - & - & - & - & - & -\\ \cline{1-14}

    \multicolumn{1}{c|}{\multirow{6}{*}{s0}} & GA-DDPG~\cite{wang2022goal}  & 50.00 &  \textbf{7.14} & 22.5 & 36.81 & \textbf{4.66} & 23.6 & 23.59 & 7.31 & 10.3 & 46.7 & \textbf{5.50} & 26.9\\
    
    \multicolumn{1}{c|}{\multirow{2}{*}{train on}}   &Handover-Sim2real~\cite{christen2023learning}& 75.23 &    7.74  &  \textbf{30.4}  &  68.75  & 6.23 & 35.8 & 29.17 & 6.29 & 15.0 & 52.40 & 7.09 & 23.8\\
    
    \multicolumn{1}{c|}{}  &\textcolor{blue}{Handover-Sim2real}\red{*}~\cite{christen2023learning} & 64.35 &  7.61  &  26.7  &  25.69  & 5.43 & 15.0 & 28.56 & 4.73 & 17.9 & 30.60 & 5.98 & 16.5\\

    \cline{2-14}
    
    \multicolumn{1}{c|}{} &  Destination Planning   &  74.31 &  9.01 & 22.8 & 76.16 & 6.98 & 35.2 & 25.68 & 5.96 & 14.1 & 48.4 & 8.94 & 15.1\\
    
    \multicolumn{1}{c|}{} & Dense Planning & 74.77 & 9.54 & 19.8 & 75.45 & 7.32 &  33.0 &  27.30 & 6.26 & 14.1 & 52.3 & 9.24 & 15.1 \\

    \multicolumn{1}{c|}{} & Landmark Planning & 77.78 & 9.24 & 22.3 & 79.17 & 7.26 & 34.9 & 29.63 & 6.23 & 15.4 & 54.2 & 9.02 & 16.6\\  
    \cline{1-14}
    \cline{1-14}

    \multicolumn{1}{c|}{\multirow{6}{*}{t0}} & {GA-DDPG}~\cite{wang2022goal} & 54.76 & 7.26 & 24.2  & 44.68 & 5.30 & 26.5  & 24.05 & 4.70 & 15.3  & 25.50 & 5.86 & 14.1\\
    
    \multicolumn{1}{c|}{\multirow{2}{*}{train on}}  & {Handover-Sim2real}~\cite{christen2023learning}    &  65.97 &  7.18  & 29.5 & 62.50 & 6.04 & 33.5 & 33.71 & 5.91 & 18.4 & 47.10 & 6.35 & 24.1\\

    \multicolumn{1}{c|}{}  & {\textcolor{blue}{Handover-Sim2real}\textcolor{red}{*}}~\cite{christen2023learning}    & 63.55  &  7.58 & 26.5 & 38.89 & 5.29 & 23.1 & 33.31 & \textbf{4.64} & 21.4 & 33.35 & 5.81 & 18.4 \\
    \cline{2-14}

    \multicolumn{1}{c|}{}  & Destination Planning   & 0.93 &  12.80 & 0.01 & 6.48 & 12.41 & 0.3 & 5.96 & 8.81 & 1.9 & 1.60 & 12.03 & 0.1\\
    
    
    \multicolumn{1}{c|}{} & Dense Planning  & 81.48 & 9.51 & 21.9 & 84.95 & 7.45 & 36.3 & 38.04 & 7.16 & 17.1 
    & 57.90 &  8.85 & 18.4\\

    \multicolumn{1}{c|}{} & Landmark Planning & \textbf{86.57} & 8.81 & 28.0 & \textbf{85.65}  & 6.58 & \textbf{42.8} & \textbf{41.43} &  6.01 &\textbf{22.3} &\textbf{68.33} &  7.70 & \textbf{27.9}\\
    
    
    \cline{1-14}
    \end{tabularx}
     \caption{\textbf{Evaluating on different benchmarks. } We compare our method against baselines from the test set of HandoverSim \cite{chao2022handoversim} benchmark (``s0 (sequential)'' and ``s0 (simultaneous)'') and our \simabbns~benchmark (``t0'' and ``t1''). We use the best-pretrained models from the repositories of GA-DDPG~\cite{wang2022goal} and Handover-Sim2real~\cite{christen2023learning} for evaluation. The results for our method are averaged across 3 random seeds. Note that S means success rate(\%). T means time(s). AS means average success(\%) as defined in Equation \ref{equ:ap}. $\dagger$: This method \cite{wang:rss2020} is evaluated with ground-truth states and cannot handle dynamic handover like ``s0 (Simultaneous)'', ``t0'' and ``t1''.\textcolor{red}{*}: \revise{We reproduce the results of HandoverSim2real in the true simultaneous setting as detailed in Section \ref{exp:benchmark} to make a fair comparison.}
    }
     \label{tab:main_exp_supp}
    \end{table*}

\subsubsection{Trajectory Resampling Strategy}
In this section, we introduce another resampling strategy designed to enhance the vision-action correlation in expert demonstrations.

Upon successfully solving the Inverse Kinematics (IK) to achieve the 6D target grasping pose, the subsequent task involves generating a trajectory from the initial 7D joint configuration \(C\) to the destination 7D joint configuration \(D\). Through optimization, OMG-Planner \cite{wang:rss2020} produces a trajectory \(C_0=C,C_1,\cdots,C_{N-1},C_N=D\) with fixed time steps \(N\). However, a challenge arises as the step size of the expert action is influenced by the distance between the initial end effector pose and the target grasping pose. Specifically, if the target grasping pose is far away from the initial end effector pose, the expert will move faster; conversely, if the target grasping pose is close to the initial end effector pose, the expert will move slower. This variability in step size can potentially confuse the vision-based model, which lacks awareness of the initial end effector pose and struggles to discern the expert's speed. 

To address this issue, we conduct a resampling process to refine the obtained trajectory. Let \(s_i=\sum_{j=1}^{i}\|C_j-C_{j-1}\|\) represent the accumulated step length, and \(L\) denote the hyperparameter controlling the desired step length. The resampled trajectory, denoted as \(C'_0=C_0,C'_1,\cdots,C'_{M-1},C'_M=C_N\), where \((M-1)L< s_N \le ML\), and for each \(1\le i\le M-1\), suppose \(s_j\le iL\le s_{j+1}\), then 
\begin{equation}
C'_i=\frac{s_{j+1}-iL}{s_{j+1}-s_j}C_j+\frac{iL-s_j}{s_{j+1}-s_j}C_{j+1}.
\end{equation}

The resampling ensures that, regardless of the proximity between the initial end effector pose and the target grasping pose, the resulting trajectory maintains a consistent step length. This characteristic makes the expert demonstrations more conducive to distillation for vision-based models.

\subsection{Forecast-Aided 4D Imitation Learning}



Figure~\ref{fig:flow_pipeline} provides an overview of our Forecast-Aided 4D Imitation Learning process. Similar to Handover-Sim2real~\cite{christen2023learning}, We initiate the pipeline by obtaining an egocentric hand and object point cloud from the simulator, together with the one-hot encoding for hand/object labels. We augment the feature of each point with its 3D coordinates in the past \(n\) time steps, computed from the estimated flow information and the current 3D coordinates. As a result, each point has a feature vector of length $3+2+3\cdot n$.

We then introduce the specific method for computing flow information. Given the end effector pose, we convert the point cloud from the egocentric frame to the static world frame and store it in a buffer. In each time step, we retrieve the point clouds of several past time steps and leverage the Iterative Closest Point (ICP) registration algorithm~\cite{rusinkiewicz2001efficient} to estimate transformation matrices between the current point cloud and past point clouds in the world frame. While these transformations may be slightly imprecise due to the incomplete point cloud input, they can provide sufficient flow information for each point. Finally, the flow information is converted back to the current egocentric frame, which serves as an important part of our feature representation.

Then we feed the point cloud with processed features into PointNet++~\cite{qi2017pointnet++} to obtain a global low-dimensional representation. This representation is then decoded by two heads: the policy head and the prediction head. The policy head decodes it into a 6D egocentric action, and $\mathcal{L}_{action}$ is computed following the approach defined in ~\cite{li2018deepim}. Simultaneously, the prediction head decodes the representation into transformations between the current and future object poses, with $\mathcal{L}_{pred}$ computed as a motion prediction loss.

Similar to Handover-Sim2real~\cite{christen2023learning}, our policy only outputs 6D egocentric actions in a closed loop. For decisions on whether to grasp the object and place it in the target location, we adopt a heuristic method akin to GA-DDPG~\cite{wang2022goal}. Specifically, if the number of points in the gripper's vicinity exceeds a predefined threshold, the robot attempts to grasp the object and retract to the target location in an open-loop fashion, foregoing the execution of the egocentric actions predicted by the policy network.

\section{Simulation Experiments Details}
\label{sec:B}

\subsection{\revise{Discussions on the Simultaneous Setting}}
It's essential to clarify that in handoverSim~\cite{chao2022handoversim} and Handover-Sim2real~\cite{christen2023learning}, the simultaneous setting (also referred to as ``w/o hold'') implies that the robot is allowed to move from the beginning of the episode. \textbf{We adhere to this definition in our GenH2R-Sim and our methods}. In the settings ``s0 (Simultaneous)'', ``t0'', and ``t1'', the robot initiates movement immediately upon detecting the object.

However, we observed that in the code of Handover-Sim2real, a parameter named ``TIME\_WAIT'' is used to specify the time to wait before executing the actual action in different settings. In ``s0 (Sequential)'', ``TIME\_WAIT'' is set to 3s (matching the 3-second duration of DexYCB~\cite{chao2021dexycb} handover motion), but in ``s0 (Simultaneous)'', ``TIME\_WAIT'' is set to 1.5s, which implies a 1.5s wait for the simultaneous setting. We believe this value should be 0s for the simultaneous setting. 

The author adjusted this parameter after observing that humans typically move faster than the robot. This modification aims to prevent collisions, especially when the robot approaches the human while the human is also approaching the object. The change is implemented to reduce the number of failures caused by attempting to grasp while the human is still in motion or before the human completes picking up the object from the table.

We believe that improving performance makes sense, but we consider the true simultaneous setting to be closer to real-world scenarios. It's crucial not to make the person wait for an extended period during a short-term handover process, so we avoid adjusting the waiting time manually.

For a fair comparison, we reproduce the results in the true simultaneous setting.

As highlighted in the blue rows of Table~\ref{tab:main_exp_supp}, our method demonstrates significant improvements. 
When trained on  ``s0'', our method achieves improvements of  13.43\%, 53.48\%, 1.07\%, and 23.6\% in the successful rate of ``s0 (Sequential)'', ``s0 (Simultaneous)'', ``t0'', and ``t1'' settings, respectively. When trained on  ``t0'', our method achieves improvements of 23.02\%, 46.76\%, 8.12\%, and 34.98\% across the four test sets. Notably, our method excels in the simultaneous setting when hands are in motion. This highlights that our distillation-friendly demonstrations can better extract valuable insights from more complex scenarios and showcase enhanced generalizability compared to the baseline.

\subsection{Baseline Experiments}

When training Handover-Sim2real, we follow the two-stage teacher-student training approach outlined in the paper.
In the pretraining stage, the duration of hand and object movement is clipped to 3 seconds, and the expert waits for 3 seconds before planning to grasp the static object. In the finetuning stage, the entire movement is used.  In the original codebase, the robot waits 1.5 seconds. In our reproduced version, the robot does not need to wait, moving directly with the dynamic hand and object.



\subsection{More Ablations}

We conducted additional ablations for our method, as shown in Table~\ref{tab:more_ablation}. All these methods are trained in ``t0'' and tested in ``t0''.

In Section \ref{subsec:4dnetwork} of the manuscript, we mentioned that frame stacking is a straightforward approach but struggles to capture both motion and geometry effectively. To quantitatively demonstrate this, we compared it with our method based on forecast-aided 4D imitation learning and found a 5.26\% decrease in the success rate. This highlights the effectiveness of our method in learning from 4D spatial-temporal information.

\revise{Moreover, excluding the endpoints from consideration when selecting the landmark results in a 1.70\% decrease in the success rate. This indicates the importance of incorporating the endpoints when choosing the landmark state.}

\begin{table}[t]
\centering
\setlength{\tabcolsep}{3mm}
\small
\begin{tabular}{cccc}
\hline
\textbf{Methods}              & S & T &  AS\\ \hline

w/o Flow            &     ~31.66~        &   \textbf{~5.67~}   &  ~17.9~\\ 
w/o Prediction       &  39.18     &     6.11        &      20.7\\
w/o Flow \& Prediction   &    37.04       &   5.93 &     20.1\\ 

\textcolor{blue}{w/o Endpoints} &     39.73        &     5.90   & 21.7\\ 
\textcolor{blue}{Frame Stacking} & 35.17 & 5.82 & 19.4 \\
Ours                 &       \textbf{41.43}      &    6.01  &   \textbf{22.3}\\ \hline


\end{tabular}
 \caption{\textbf{Ablations on different modules.} ``w/o Flow'' means not using flow information in the input. ``w/o Prediction'' means not adding prediction loss in the final loss. 
}
 \label{tab:more_ablation}
\end{table}



\begin{figure}[!t]
    \centering
    \includegraphics[width=\linewidth]{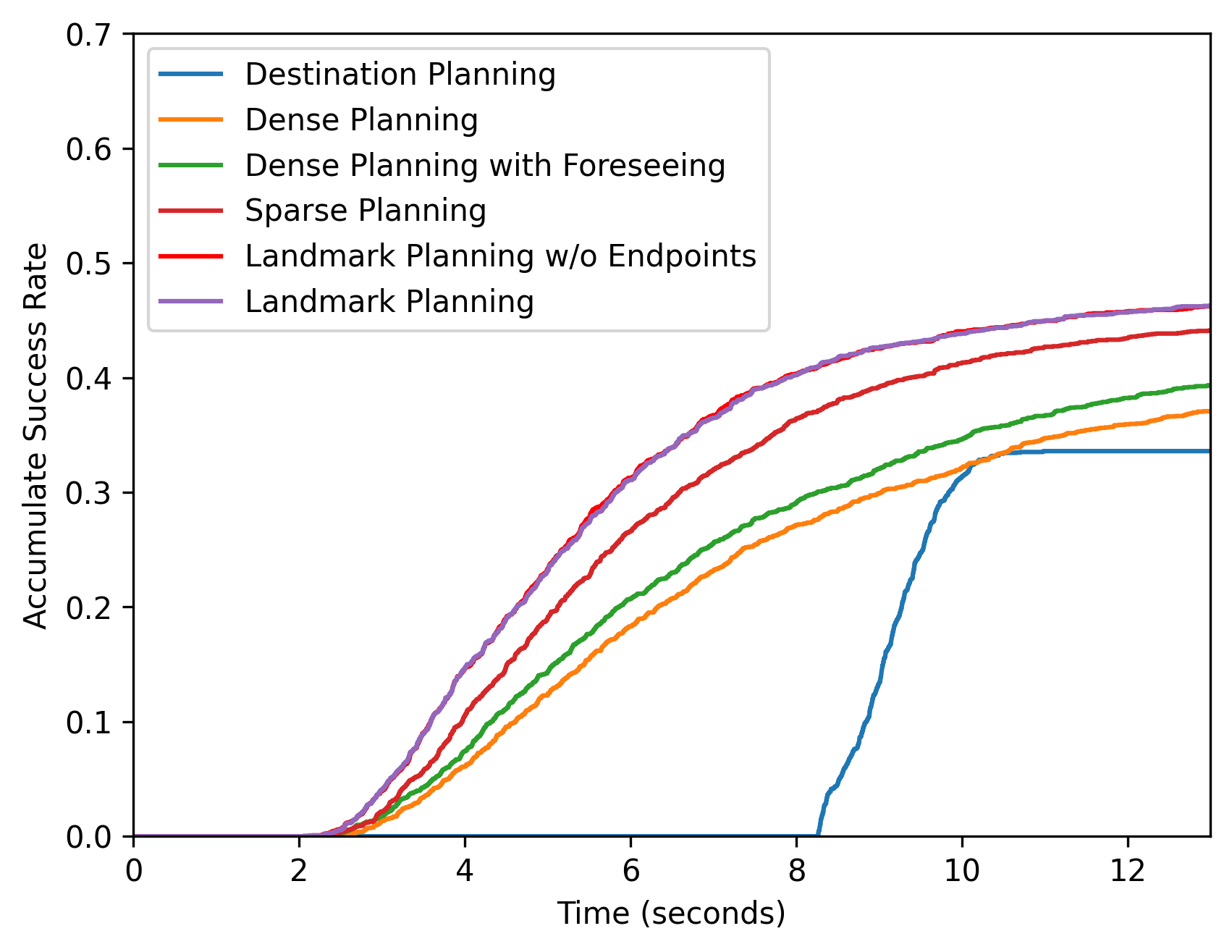}
    \caption{\textbf{Comaprison of different expert demonstrations.}}
    \label{fig:demo}
\end{figure}

\subsection{Evaluating on Demonstration Generation}

Figure \ref{fig:demo} compares different expert demonstration generation variants by showing their accumulated success rate \wrt to success time on ``t0''.

Destination Planning is suboptimal, as it plans a straight trajectory directly to the destination, which is not only slow but also nearly impossible to be distilled to vision-based agents when the object trajectory is complex.

With access to future object states and with less frequent replanning periods, Landmark Planning exhibits a larger success rate and faster success time compared with Dense Planning. To ablate the two factors, we also analyze two additional settings. In Dense Planning with Foreseeing, OMG-Planner still always replans at each step, but based on the future object states. In sparse Planning, OMG-Planner replans with the same sparsity as Landmark Planning, but it can only plan based on the current object state. We can analyze from the curves that both decreasing the replanning period and foreseeing future states can help the expert to achieve a higher success rate and lower success time.

\subsection{Training Details}
In our training process, we employ a single Nvidia GeForce RTX 3090 (24GB) with a batch size of 256. The training spans 80,000 iterations, with each iteration involving the random sampling of 256 observation-action pairs from demonstrations. We use the Adam optimizer with a learning rate of 0.001 and weight decay of 0.0001. The entire process takes approximately 8 hours to train our method. We incorporate flow information from the last 3 time steps and calculate the prediction loss for the next 3 time steps with the weighting hyper-parameter $\lambda$ as 0.1.

\section{Real World Experiments Details}
\label{sec:C}
\subsection{Setup}

\begin{table*}[t]
\centering
\small
\begin{tabular}{c|ccc|ccc}
\hline
\multirow{2}{*}{} & \multicolumn{3}{c|}{Simple setting}                       & \multicolumn{3}{c}{Complex setting}                       \\ \cline{2-7} 
                  & GA-DDPG\cite{wang2022goal}           & Handover-Sim2real\cite{christen2023learning} & Ours              & GA-DDPG\cite{wang2022goal}           & Handover-Sim2real\cite{christen2023learning} & Ours              \\ \hline
1. plastic mug    & 5 / 6             & 4 / 6             & 6 / 6             & 2 / 6             & 2 / 6             & 4 / 6             \\ \hline
2. EFES bottle    & 5 / 6             & 4 / 6             & 6 / 6             & 4 / 6             & 3 / 6             & 5 / 6             \\ \hline
3. Cheez-It box   & 1 / 6             & 4 / 6             & 4 / 6             & 3 / 6             & 1 / 6             & 4 / 6             \\ \hline
4. sticky tape    & 3 / 6             & 3 / 6             & 5 / 6             & 3 / 6             & 3 / 6             & 4 / 6             \\ \hline
5. Pringle can    & 4 / 6             & 2 / 6             & 6 / 6             & 1 / 6             & 1 / 6             & 4 / 6             \\ \hline
total             & 18 / 30 (60\%) & 17 / 30 (57\%) & \textbf{27 / 30 (90\%)} & 13 / 30 (43\%) & 10 / 30 (33\%) & \textbf{21 / 30 (70\%)} \\ \hline
\end{tabular}

 \caption{\textbf{User study for sim-to-real experiments.} Each method was evaluated by six individuals for every object in both the simple and complex settings. Failure scenarios included collisions with the human hand, dropping to the table, or exceeding the time limit ($T_{\text{max}} = 13$ seconds). Our method consistently outperformed the baselines in the real-world handover system in both simple and complex settings, aligning with the results observed in the simulation experiments.
}
 \label{tab:real_supp}
 
\end{table*}
\begin{table}[htbp]
\centering
\setlength{\tabcolsep}{3mm}
\scriptsize
\begin{tabular}{p{0.16\textwidth}cccc}
\hline
\textbf{Objects}              & GA-DDPG & Handover-Sim2real &  Ours\\ \hline
1. Transparent bottle       &  1 / 6     &     2 / 6        &      \textbf{3 / 6}\\
2. Transparent beaker       &  1 / 6     &     0 / 6        &      \textbf{2 / 6}\\
3. Transparent cup       &  1 / 6     &     2 / 6        &      \textbf{4 / 6}\\
4. Regular cup       &  \textbf{5 / 6}     &     1 / 6        &      \textbf{5 / 6}\\
5. Glue stick       &  3 / 6     &     2 / 6        &      \textbf{6 / 6}\\
6. Large teacup       &  3 / 6     &     2 / 6        &      \textbf{4 / 6}\\
7. Blue ball      &  1 / 6     &     1 / 6        &      \textbf{4 / 6}\\
8. Small sponge     &  3 / 6     &     1 / 6        &      \textbf{6 / 6}\\
9. Tape measure       &  3 / 6     &    \textbf{ 4 / 6 }       &      \textbf{4 / 6}\\
10. Large bucket       &  4 / 6     &     1 / 6        &      \textbf{5 / 6}\\
11. Stapler       & \textbf{ 3 / 6}     &     2 / 6        &      \textbf{3 / 6}\\
12. Disinfectant       &  3 / 6     &     2 / 6        &      \textbf{4 / 6}\\
13. Packaging box       &  2 / 6     &     2 / 6        &      \textbf{4 / 6}\\
14. Saw       &  2 / 6     &     2 / 6        &      \textbf{5 / 6}\\
15. Book       &  4 / 6     &     2 / 6        &      \textbf{6 / 6}\\  \hline
Overall       &   43.3\%     &    28.9\%     &    \textbf{72.2\%} \\ \hline
\end{tabular}
 \caption{\textbf{User study for sim-to-real-experiments in various and novel objects.} Each method was evaluated by six individuals for every object. Our method outperformed the baselines by a large margin.}
 \vspace{-4mm}
 \label{tab:real_supp2}
\end{table}

\begin{figure}[t]
  \centering
  \includegraphics[width=\columnwidth]{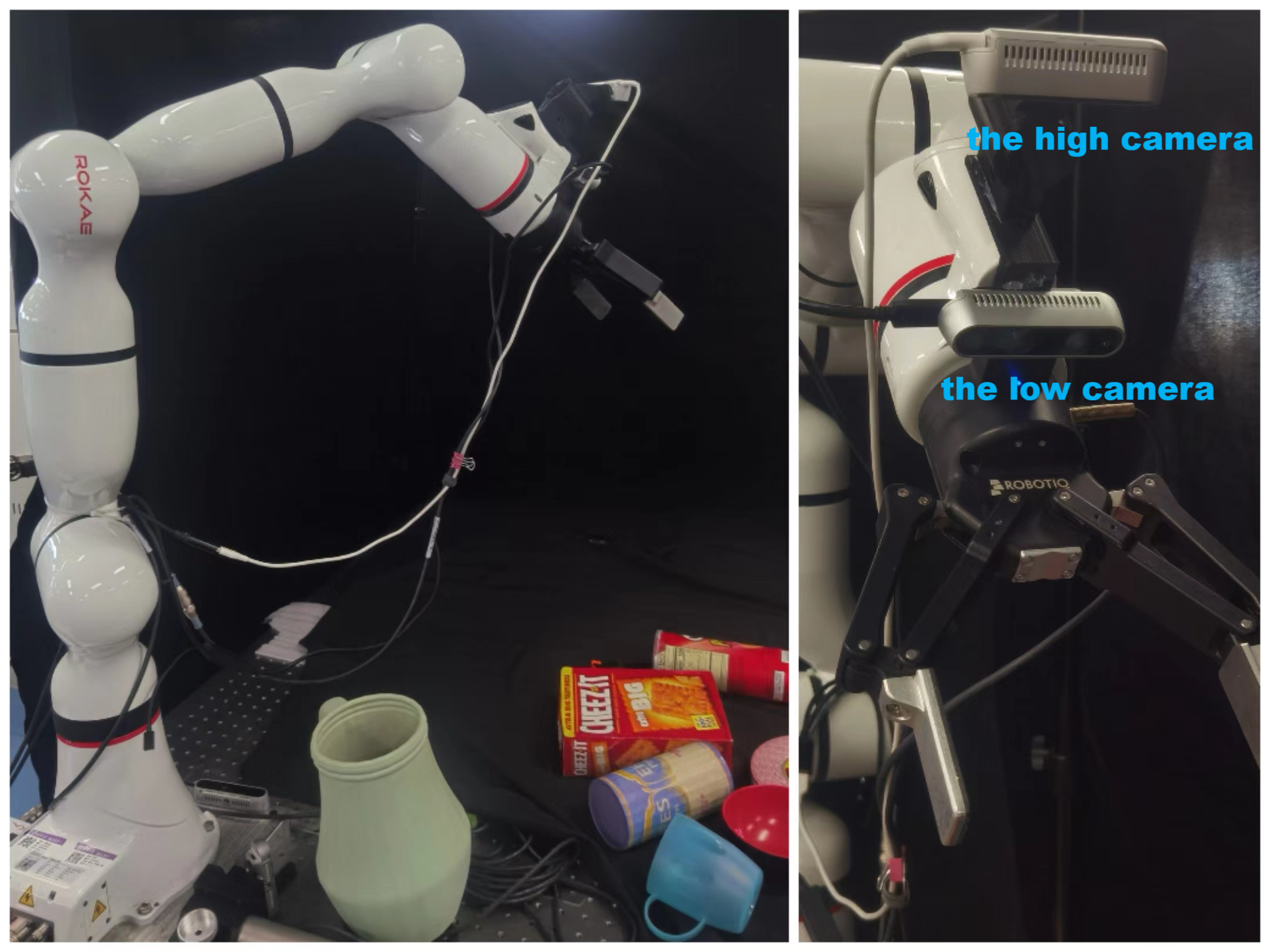}
  \caption{\textbf{Real-world Handover System Setup.} Our system consists of an xMate 3 robot, which is similar to a Franka Panda robot, and two RealSense Depth Camera D435 devices.}
\label{fig:real_setup}
\vspace{-0.4cm}
\end{figure}

\begin{figure}[t]
  \centering
  \includegraphics[width=\columnwidth]{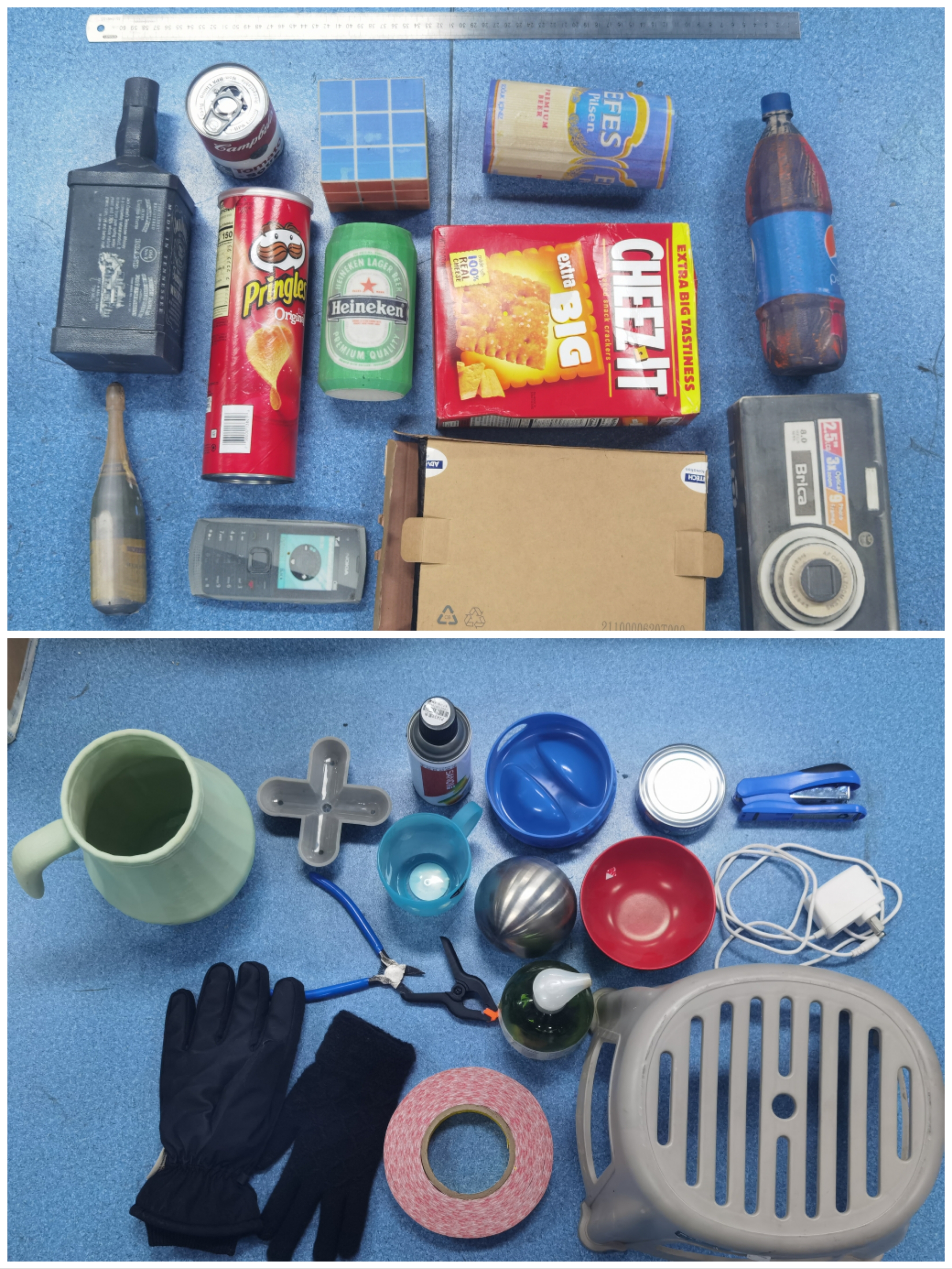}
  \caption{\textbf{Various objects for real-world handover.} The image above displays relatively simple objects for handover, such as the can, the box, the bottle, or some square objects. In contrast, the image below showcases more challenging objects for handover, including the plastic stool, the teapot, the sticky tape, or some soft objects with diverse shapes.}
\label{fig:different_task}
\vspace{-0.4cm}
\end{figure}
In our real-world handover system, illustrated in Figure~\ref{fig:real_setup}, we utilize a ROKAE xMate 3 ER series flexible collaborative robot along with two Intel RealSense Depth Camera D435 devices.


Since the action space is 6D Cartesian, while our policies and the baselines are trained with Franka Panda robot in HandoverSim~\cite{chao2022handoversim} and \simabbns, we can deploy them to the xMate3 robot despite their morphology difference, as they have own position controllers. Thus, our real-world experiments not only assess the policy's generalization for sim-to-real transfer but also evaluate its adaptability to different robots.

Our real-world handover system incorporates two Intel RealSense Depth Camera D435 devices to enhance the egocentric point cloud. In Figure~\ref{fig:real_setup}, the higher camera captures the point cloud in proximity to the robot gripper but lacks visibility further ahead. Conversely, the lower camera captures the point cloud ahead of the robotic arm but misses details near the gripper. By merging the point clouds from both cameras, we achieve a comprehensive view, which is beneficial for the deployed policies.

The robot initiates movement upon perceiving a point cloud. In case the object is not visible during the handover process, the robot tracks the object's last known pose. The baselines, GA-DDPG~\cite{wang2022goal} and Handover-Sim2real~\cite{christen2023learning}, are treated similarly to~\cite{christen2023learning}. For GA-DDPG, the pre-trained policy model is loaded, and heuristic methods determine whether to grasp. For Handover-Sim2real, both the pre-trained policy model and the pre-trained grasp prediction network are loaded.

\subsection{User Study}
The study involved 6 participants who compared our forecast-aided 4D imitation learning method based on landmark planning demonstrations, with two baseline methods, GA-DDPG and Handover-Sim2real. The first user study contains 5 different objects in two settings. The second user study contains 15 various and novel objects.


For the first user study, The selected 5 objects for evaluation include a mug, a bottle, a cracker box, a sticky tape, and a chips can. The cracker box is an object shown in DexYCB~\cite{chao2021dexycb} trajectories, while the other 4 novel objects may exhibit more diverse geometries.

In the simple setting, users hand each object to the gripper in a straightforward manner. In the complex setting, users execute a relatively long and quick trajectory, involving both translations and rotations. 
For each specific object, we try to ensure that each participant executed a similar trajectory in the same setting for different methods, ensuring a fair comparison.

Table \ref{tab:real_supp} provides a detailed breakdown of the results presented in Table 3 of the manuscript. Our method is compared with baselines across different settings, revealing a remarkable 34\% improvement in the simple setting and a substantial 40\% improvement in the complex setting from Handover-Sim2real. Notably, in the simple setting, our method demonstrates great generalizability to various objects, including new objects with different geometries or similar objects of different sizes. In the complex setting, our method exhibits smooth object tracking with predictive intention, resembling a more human-like approach to grasping handed objects. Further analysis will be elaborated in our accompanying video. It is noteworthy that Handover-Sim2real exhibits a lower success rate compared with GA-DDPG. One possible explanation is that the pre-trained grasp prediction network may not be as robust as heuristic methods in determining whether to grasp, which may not be able to generalize well to novel objects and potentially increase the sim-to-real gap.

For the second user study, we also expanded the real-world experiments comparing our method and baselines. 6 users participated by handing over \textbf{15 diverse and novel objects} varying in geometry, size, and transparency. Users provide various trajectories, occasionally adopting a less cooperative or adversarial manner. They blindly test each policy on identical trajectories. Our method consistently outperforms baselines as shown in Table \ref{tab:real_supp2}.
While adversarial behavior and weird object geometry lead to failures in baselines, our approach generalizes well and adeptly adjusts to the hand trajectory quickly.
For failure cases, our method faces challenges with transparent objects due to corrupted depth. The final version will include detailed tests and elaborations.

\subsection{Generalization Study}




In addition to direct comparisons with baseline methods, we conducted numerous real-world handover experiments involving different trajectories and objects.

Figure~\ref{fig:different_task} showcases two sets of objects used in our experiments. The simple set comprises regular objects similar to those used in DexYCB~\cite{chao2021dexycb} or HandoverSim~\cite{chao2022handoversim}, which are easier to pass and grasp. The difficult set includes more challenging objects with diverse shapes and geometries. We introduced variations in human behavior, such as different grasping poses or handover trajectories.

\begin{figure*}[t]
  \centering
  \includegraphics[width=\linewidth]{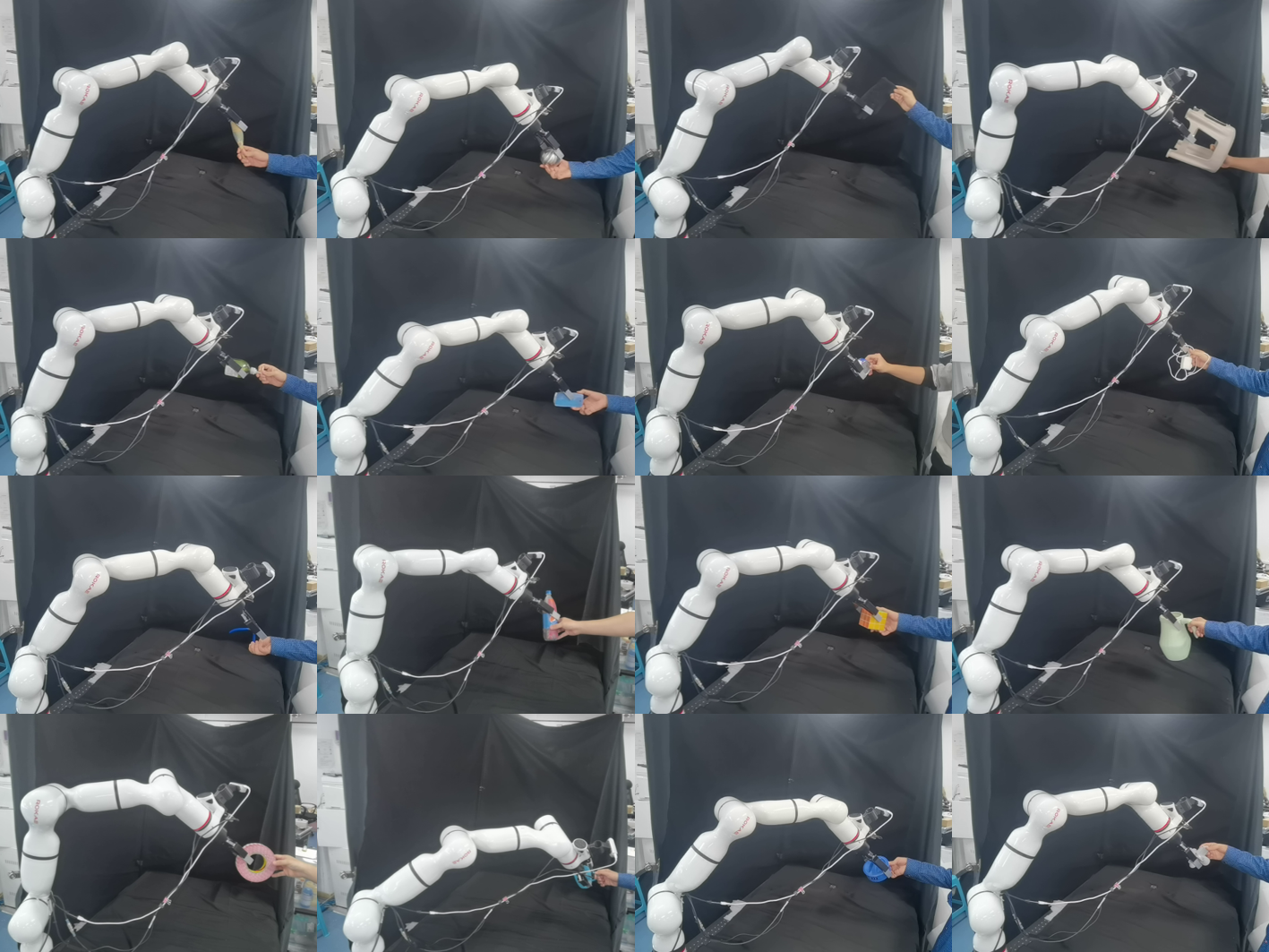}
  \caption{\textbf{Qualitative real-world results with various objects}. In the real-robot system, we qualitatively assess the generalization ability of our method by testing it with various objects. }
\label{fig:real_exp_with_various_objects}
\end{figure*}

As shown in Figure \ref{fig:real_exp_with_various_objects}, our qualitative experiments reveal the robustness of our policy, crafted through extensive large-scale demonstrations and an imitation learning framework. Trained within the \simabb environment, our policy showcases effective generalization across diverse objects and various handover scenarios. 

\section{Limitations and Future Work}
\label{sec:D}
While in this paper significant progress has been achieved in the H2R handover task, we acknowledge certain limitations that could serve as inspiration for exciting future research.

In aspects of robot morphology, we concentrate on the relatively simple 7DoF robotic arms, characterized by a confined activity region and limited motion capabilities. In contrast, robots equipped with a movable base exhibit a broader range of motion, enabling them to navigate and interact within a more extensive spatial environment, enhancing their versatility and efficacy in various human-robot interaction tasks.

In aspects of human modeling, our current focus on the object and hand poses neglects the consideration of the entire human body. In real-world scenarios, robots may need to take into account not only the hand pose and trajectories but also the motion of the entire human body for more dynamic and generalizable interactions. Extending the simulation environment to model a more complex representation of the human, including body movements, poses a challenging yet practical avenue for future work in policy learning.

In aspects of human intention, our simulator currently does not incorporate human intention. Existing simulation environments have mainly focused on physical modeling, lacking representation of human behavior. In HandoverSim~\cite{chao2022handoversim}, for instance, the human hand does not respond to the robot's actions. In \simabbns, we introduce a more interactive element, where the human hand stops moving and waits for handover when the robot arm is close to the object. However, there is room for more complex and interesting modeling of human behavior. For instance, when the gripper moves rapidly toward the hand, the human may perceive danger and retract the hand. Introducing more sophisticated representations of human behavior in the simulator is crucial for a human-centric handover process.

\end{document}


\title{Semantic Complete Scene Forecasting from a 4D Dynamic Point Cloud Sequence —  Supplementary Material}

\author{First Author\\
Institution1\\
Institution1 address\\
{\tt\small firstauthor@i1.org}
\and
Second Author\\
Institution2\\
First line of institution2 address\\
{\tt\small secondauthor@i2.org}
}

\twocolumn[{%
\maketitle
}]

This document provides a list of supplemental materials to support the main paper.

\begin{itemize}
        \item {
            \textbf{Term Correction. }In Experiments Section, we misuse the terms \textit{Prediction} and \textit{Forecasting}, which may lead to confusion when reading our paper. We will point out each line that got them wrong in \Cref{term}.
        }
        \item {
            \textbf{Additional Ablation Studies. }We provide additional
ablation studies from different factors in \Cref{ablation}. Specifically, we examine the correlation between skip connections and dilated convolution layers, and compare our 4D point encoder with 4D voxel encoder.
        }
        \item {
            \textbf{Dataset Description. }We detail the generation process of IGPLAY and IGNAV in  \Cref{dataset_desc}. Multi-type data formats allow us to conduct a variety of experiments related to the SCSF task.
        }

        \item {
            \textbf{Additional Results. }We provide additional experiment results. Also, we provide visualizations in a better way. in\Cref{additional_results}. 
        }

        \item {
            \textbf{Baseline Details. }In order to fit other relevant models to the new SCSF task, we need to do some modifications. We will give details about this process and  provide more baseline experiments  in\Cref{baseline_detail}. 
        }
        
        \item {
            \textbf{Semantic Scene Completion with SCSFNet. }Ignoring future forecasting, we can degenerate  SCSFNet to tackle SSC(Semantic Scene Completion) problem. We present how to solve the SSC problem for IGPLAY, IGNAV, and NYUCAD in the form of 3D or 4D input in \Cref{4DSSC}.
        }
        
    \end{itemize}

\input{supplementary/term.tex}
\input{supplementary/addition_ablation.tex}
\input{supplementary/dataset_sesc.tex}
\input{supplementary/additional_results.tex}
\input{supplementary/baseline.tex}
\input{supplementary/ssc.tex}

{\small
\bibliographystyle{ieee_fullname}
\bibliography{egbib_sup}
}